\documentclass[runningheads]{llncs}

\usepackage[T1]{fontenc}

\usepackage{graphicx}
\usepackage{booktabs}
\usepackage{multirow}
\usepackage{svg}
\usepackage{changepage}
\usepackage{color}
\usepackage{orcidlink}
\usepackage{array}
\usepackage{changepage}

\begin{document}

\title{A Weak Supervision Approach for Monitoring Recreational Drug Use Effects in Social Media}

\titlerunning{Weak Supervision for Drug Effects on Social Media}

\author{Lucía Prieto-Santamaría\inst{1,2}\orcidlink{0000-0003-1545-3515} \and Alba Cortés Iglesias\inst{1} \and
 Claudio Vidal Giné\inst{3}\orcidlink{0000-0002-4936-007X} \and Fermín Fernández Calderón\inst{4,5}\orcidlink{0000-0002-2981-1670} \and Óscar M. Lozano\inst{4,5}\orcidlink{0000-0003-2722-6563} \and
Alejandro Rodríguez-González\inst{1,2}\orcidlink{0000-0001-8801-4762}}

\authorrunning{L. Prieto-Santamaría et al.}

\institute{Escuela Técnica Superior de Ingenieros Informáticos, Universidad Politécnica de Madrid, Spain \\
\email{lucia.prieto.santamaria@upm.es}\\ \email{alba.cortes@alumnos.upm.es}\\
\email{alejandro.rg@upm.es}\\
\and
Centro de Tecnología Biomédica, Universidad Politécnica de Madrid, Spain\\
\and
Asociación Bienestar y Desarrollo, Energy Control, Spain\\
\email{claudiovidal@energycontrol.org}
\and
Clinical and Experimental Psychology Department, University of Huelva, Spain\\
\email{fermin.fernandez@dpces.uhu.es}\\ \email{oscar.lozano@dpsi.uhu.es}
\and
Research Center for Natural Resources, Health and the Environment, University of Huelva, Spain
}

\maketitle              % typeset the header of the contribution
\begin{abstract}
Understanding the real-world effects of recreational drug use remains a critical challenge in public health and biomedical research, especially as traditional surveillance systems often underrepresent user experiences. In this study, we leverage social media (specifically Twitter) as a rich and unfiltered source of user-reported effects associated with three emerging psychoactive substances: ecstasy, GHB, and 2C-B. By combining a curated list of slang terms with biomedical concept extraction via MetaMap, we identified and weakly annotated over 92,000 tweets mentioning these substances. Each tweet was labeled with a polarity reflecting whether it reported a positive or negative effect, following an expert-guided heuristic process. We then performed descriptive and comparative analyses of the reported phenotypic outcomes across substances and trained multiple machine learning classifiers to predict polarity from tweet content, accounting for strong class imbalance using techniques such as cost-sensitive learning and synthetic oversampling. The top performance on the test set was obtained from eXtreme Gradient Boosting with cost-sensitive learning (F1 = 0.885, AUPRC = 0.934). Our findings reveal that Twitter enables the detection of substance-specific phenotypic effects, and that polarity classification models can support real-time pharmacovigilance and drug effect characterization with high accuracy.

\end{abstract}
\section{Introduction}

Recreational drug use represents a persistent and complex public health challenge, contributing to both acute and chronic adverse health outcomes. Substances such as: ecstasy, GHB, and 2C-B are widely consumed for their psychoactive properties, often outside regulated medical contexts~\cite{teter_comprehensive_2001,gahlinger_club_2004,schifano_novel_2015}. Understanding the effects these substances produce in real-world settings is essential to evaluate their risks, identify patterns of misuse, and detect emergent phenomena. Traditional pharmacovigilance systems, while indispensable, rely heavily on structured reporting from healthcare providers and institutions. These systems tend to underreport mild, subjective, or socially stigmatized experiences, and often lag behind actual behavioral trends~\cite{ricaurte_recognition_2005}.

In contrast, social media platforms, particularly Twitter (now X), offer a complementary and dynamic source of health-related data~\cite{ruths_social_2014,ji_twitter_2015,meng_national_2017,rao_digital_2024}. Users frequently share their experiences with drugs in informal language, including detailed descriptions of sensations, side effects, mood changes, and behavioral outcomes~\cite{nasralah_social_2020}. These user narratives reflect what we refer to as \textit{phenotypic manifestations}: observable or self-perceived effects resulting from drug intake. From a biomedical standpoint, such phenotypes represent the interplay of pharmacological mechanisms (e.g., receptor binding, metabolic clearance) with contextual and personal factors such as co-use of other substances, emotional state, and individual physiology~\cite{plant_consequences_2010}.

These phenotypic effects can be grouped into two categories: \textit{desired effects}, which motivate consumption (e.g., euphoria, sociability, sensory enhancement), and \textit{undesired effects}, which may indicate toxicity or adverse reactions (e.g., nausea, panic, confusion). Capturing both types of responses is crucial for informal pharmacovigilance, as it allows health authorities and researchers to track real-world impact, monitor perception trends, and identify potentially therapeutic or harmful patterns of use~\cite{carabot_understanding_2023}. Several studies have explored the potential of social media mining to monitor drug discourse. Early efforts relied on keyword matching and manual inspection to identify tweets related to substance use~\cite{cameron_predose_2013}. Later works incorporated sentiment analysis, topic modeling, and supervised classification to extract thematic patterns~\cite{tassone_utilizing_2020,nasralah_social_2020,castillo-toledo_insights_2024}.

A key limitation in this scenario is class imbalance (where negative effects dominate the training data) can skew classifier performance if not carefully addressed~\cite{rodriguez-gonzalez_identifying_2020}. Another critical gap is the underuse of biomedical semantic enrichment~\cite{sarker_portable_2015}. Tools such as MetaMap allow the mapping of free-text to structured concepts in the UMLS Metathesaurus, providing standardization and interoperability with clinical vocabularies~\cite{aronson_effective_2001}. This allows informal narratives to be interpreted in relation to known adverse events, pharmacological classes, and symptom ontologies~\cite{shyu_enabling_2024}. Some studies have shown that semantic features improve the accuracy of classification models in biomedical NLP tasks~\cite{nikfarjam_pharmacovigilance_2015}.

In this work, we present a comprehensive pipeline for detecting and classifying user-reported phenotypic effects of three recreational substances (ecstasy, GHB, and 2C-B) based on Twitter data. Our methodology is guided by a data mining methodological framework and integrates biomedical knowledge, weak supervision, and Machine Learning (ML). Specifically, we have (i) developed a weak labeling strategy based on polarity annotations of both a manually curated slang lexicon and biomedical MetaMap-extracted concepts; (ii) built a semantically enriched dataset of over 6 million tweets filtered by substance-related keywords and annotated for perceived effect polarity; and (iii) trained and evaluated several ML classifiers using text embeddings and domain-specific features, while applying class imbalance mitigation techniques and class weighting. The rest of this paper is structured as follows: Section~\ref{sec:methodology} presents the methodology, including data processing and modeling. Section~\ref{sec:results} reports the experimental results and analysis. Section~\ref{sec:conclusion} ends with the conclusions and a discussion of future directions.

% This study aims to support the growing field of AI-driven informal pharmacovigilance, demonstrating how subjective, unstructured online discourse can be transformed into structured phenotypic data for downstream biomedical applications. Beyond classification, our work opens the door to modeling interindividual variability, studying co-use effects, and identifying candidate signals for risk alerts.

\section{Methodology} \label{sec:methodology}

\begin{figure}[!ht]
\begin{adjustwidth}{-3.9cm}{-3.9cm}
\centering
\includegraphics[width=1.6\textwidth]{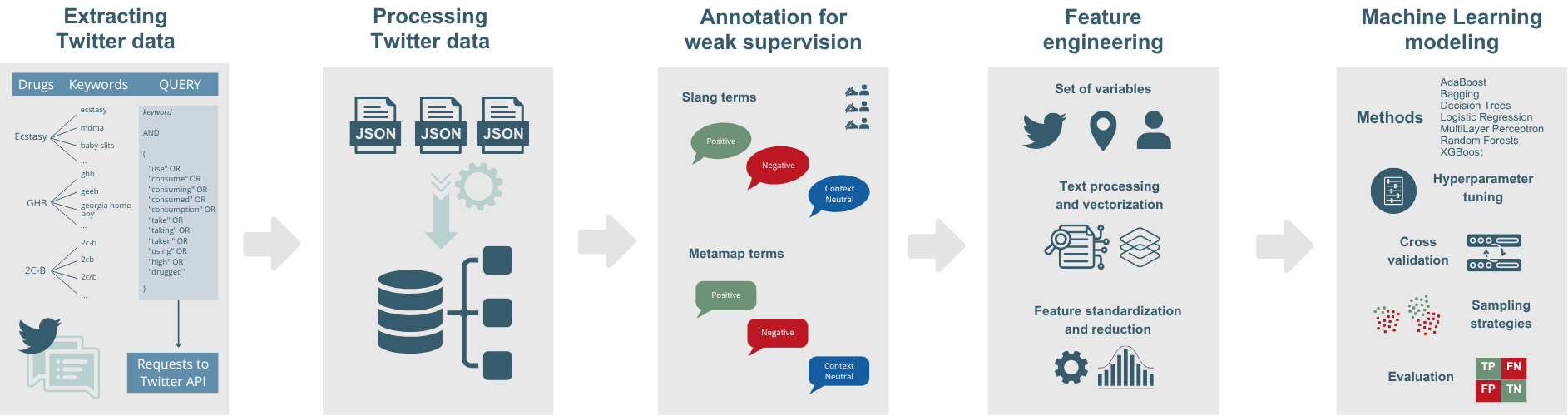}
\caption{Overview of the methodology pipeline.}
\label{fig:methodology_overview}
\end{adjustwidth}
\end{figure}

This section describes the methodological pipeline designed to collect, annotate, and analyze Twitter data related to recreational drug use. \textbf{\autoref{fig:methodology_overview}} provides an overview of the full process. We begin by explaining how Twitter data were extracted and processed (\autoref{sec:extracting_data}) to obtain structured and clean textual and metadata information. We then describe the annotation strategy based on weak supervision (\autoref{sec:annotation}) which leverages external domain knowledge to assign preliminary labels to the tweets. 
This is followed by a description of the feature engineering process to obtain a complete set of variables associated to each tweet (\autoref{sec:feature_engineering}). Finally, we present the machine learning methodology used to classify tweets according to the phenotypic effects they report related to drug use (\autoref{sec:ml_modeling}).

\subsection{Extracting and processing Twitter data}  \label{sec:extracting_data}

Firstly, we compiled tweets posted between 2010 and 2019 by accessing Twitter through its Application Programming Interface (API). 
To manage the requests, we used the Python library ``Twitter API'', which requires API keys, an endpoint, a query, and additional parameters to retrieve data. 
To identify tweets potentially mentioning the drugs under study (ecstasy, GHB, and 2C-B), we used a set of keywords associated with each drug. 
These keywords are listed in \textbf{~\autoref{tab:drug_keywords}}.

\begin{table}[!ht]
\centering
\begin{adjustwidth}{-2cm}{-2cm}
\caption{Keywords used for each drug when querying Twitter.}
\begin{tabular}{|c|c|c|}
\hline
\textbf{Ecstasy} & \textbf{GHB} & \textbf{2C-B} \\
\hline
ecstasy & ghb & 2cb \\
mdma & geeb & 2c/b \\
baby slits & georgia home boy & 2c b \\
e-bomb & liquid ecstasy & 2,5-dimethoxy-4-bromophenethylamine \\
hug drug & cherry meth & 4-bromo-2,5-dimethoxyphenethylamine \\
love drug & easy lay & afterburner bromo \\
moon rock & blue nitro & bdmpea \\
love doctor & gamma hydroxybutyrate & bromo mescaline \\
love portion \#9 & gamma-hydroxybutyric acid & cee-beetje \\
love trip & 4-hydroxybutanoic acid & erox \\
scooby snacks & date rape drug & pink cocaine \\
speed for lovers & forget-me pill & tusibi \\
xtc & forget pill & tucibi \\
3,4 methylenedioxymethamphetamine & & \\
c11h15no2 & & \\
ecstasy pill & & \\
substituted amphetamine & & \\
molly & & \\
\hline
\end{tabular}
\label{tab:drug_keywords}
\end{adjustwidth}
\end{table}

Each drug-specific query included the corresponding terms in \textbf{~\autoref{tab:drug_keywords}}, combined with the following string: \textit{AND (``use'' OR ``consume'' OR ``consuming'' OR ``consumed'' OR ``consumption'' OR ``take'' OR ``taking'' OR ``taken'' OR ``using'' OR ``high'' OR ``drugged'')}. This additional condition aimed to increase the likelihood that retrieved tweets referred to drug use rather than to unrelated contexts. Requests to the Twitter API returned data in JavaScript Object Notation (JSON) format, which included information about the tweets themselves (e.g., creation date and text), engagement metrics (number of retweets, replies, likes, and quotes), the authors who posted them, and geolocation data. Once the JSON files were retrieved through the Twitter API, the data were processed and structured into relational database tables. Tweet texts were cleaned by removing links, hashtags, and emojis, and all text was converted to lowercase. Tweet interactions were also processed to identify replies and retweets.

% The final set of stored variables included the following. 
% From each tweet, we extracted: creation date, retweet count, reply count, like count, quote count, tweet text, author ID, geolocation bounding box, country, country code, place name, and place type. 
% Regarding the authors, we collected: username, display name, verification status, follower count, following count, tweet count, listed count, and self-declared location.

\subsection{Annotation for weak supervision} \label{sec:annotation}

Since manually annotating a large collection of tweets for drug-related content is both time-consuming and costly, we adopted a weak supervision strategy to generate preliminary labels automatically. This approach leverages domain-specific knowledge sources and linguistic heuristics to infer likely labels without the need for human annotation at the tweet level.

We used two complementary sources of information to identify whether a tweet refers to the use or effects of recreational drugs: (1) a curated list of \textbf{slang terms} associated with each of the three target substances (ecstasy, GHB, and 2C-B), and (2) biomedical entities extracted using \textbf{MetaMap}~\cite{aronson_effective_2001}, a tool developed by the U.S. National Library of Medicine that maps free text to concepts in the Unified Medical Language System (UMLS)~\cite{bodenreider_unified_2004}.

Both the curated term list and the set of extracted UMLS concepts were independently reviewed by three domain experts. Each term or concept was assigned a polarity label: positive, negative, or context-dependent, based on whether its presence was likely to indicate positive or negative drug use or effects. A majority-vote strategy was used to consolidate annotations. Specifically, if at least 60\% of annotators agreed on a polarity label, that label was assigned as the final annotation. Otherwise, the term was marked as uncertain. This step ensured that only terms with sufficient inter-annotator agreement were used in the subsequent tweet-level labeling process.

Once terms and concepts had been annotated with a final polarity, tweets were heuristically labeled based on the combination of annotations for the lexical items they contained. For each tweet, the corresponding terms or concepts found (whether from the slang list or extracted by MetaMap) were mapped to numerical scores: $+1$ for positive polarity, $-1$ for negative, and $0$ for context-dependent. The scores for all elements in a tweet were aggregated, and the resulting total score determined the tweet’s polarity: tweets with a score $>0$ were labeled as positive, those with a score $<0$ as negative, and tweets scoring exactly $0$ were labeled as context-dependent.

This polarity assignment was performed independently for the slang-based and MetaMap-based information, producing two parallel tweet classification tables. To ensure high-confidence labels, we applied a final filtering step: only tweets that were present in both tables and had matching polarity assignments (either positive or negative) were retained. Tweets labeled as context-dependent in either source, or those with discordant polarity across sources (e.g., positive in slang, negative in MetaMap), were excluded.

% The resulting dataset contains tweets that simultaneously exhibit lexical cues from both informal and biomedical vocabularies and for which there was polarity agreement across sources. This conservative filtering process resulted in a reliable and semantically rich dataset, suitable for downstream machine learning modeling.

\subsection{Feature engineering} \label{sec:feature_engineering}

Once the final dataset was consolidated, we conducted a feature engineering process to prepare the data for machine learning modeling. This process combined tweet metadata, user attributes, geolocation information, and a semantic representation of textual content based on word embeddings. We extracted a \textbf{set of relevant features} from the resulting dataset (\textbf{\autoref{tab:features}}), including three main types:

\begin{itemize}
    \item \textbf{Tweet-level features:} the tweet text, presence of multimedia content, user mentions, reference to another tweet (e.g., replies or retweets), engagement metrics, and binary indicators for whether the tweet contained terms associated with each of the three substances. The target variable for classification was included with values “positive” or “negative” according to the weak supervision strategy described in \autoref{sec:annotation}. 
    \item \textbf{Geographic features:} binary indicators for the region of origin based on the country code, and a flag indicating whether the user's location field was available.
    \item \textbf{User features:} verification status, and metrics reflecting user activity and visibility, including follower count, number of followed accounts, total number of tweets, and number of public lists the user appears on.
\end{itemize}

\begin{table}[!ht]
\begin{adjustwidth}{-2cm}{-2cm}
\centering
\caption{Set of features extracted from tweets, users, and  other metadata.}
\label{tab:features}
\begin{tabular}{|p{1.7cm}|>{\scriptsize}p{2.2cm}|>{\scriptsize}p{8cm}|}
\hline
\textbf{Type} & \textbf{Feature} & \textbf{Description} \\
\hline

\multirow{11}{*}{Tweet-level} 
& \texttt{media} & Binary indicator for presence of multimedia content in the tweet. \\
& \texttt{mention} & Binary indicator for whether the tweet mentions another user. \\
& \texttt{reference} & Binary indicator for whether the tweet is a reply or retweet. \\
& \texttt{like\_count} & Number of likes received by the tweet. \\
& \texttt{retweet\_count} & Number of times the tweet was retweeted. \\
& \texttt{reply\_count} & Number of replies to the tweet. \\
& \texttt{quote\_count} & Number of quote tweets of the tweet. \\
& \texttt{mentions\_ghb} & Indicates if the tweet contains terms associated with GHB. \\
& \texttt{mentions\_ecstasy} & Indicates if the tweet contains terms associated with ecstasy. \\
& \texttt{mentions\_2cb} & Indicates if the tweet contains terms associated with 2C-B. \\
& \texttt{classification} & Weak supervision label (positive or negative). \\
\hline

\multirow{5}{*}{Geographic} 
& \texttt{is\_europe} & Tweet was posted from a country located in Europe. \\
& \texttt{is\_africa} & Tweet was posted from a country located in Africa. \\
& \texttt{is\_asia} & Tweet was posted from a country located in Asia. \\
& \texttt{is\_america} & Tweet was posted from a country located in America. \\
& \texttt{user\_location} & Binary indicator for whether the user's location is available. \\
\hline

\multirow{5}{*}{User-level} 
& \texttt{user\_verified} & Whether the user's account is verified (1 if yes, 0 otherwise). \\
& \texttt{user\_followers} & Number of followers the user has. \\
& \texttt{user\_following} & Number of accounts the user is following. \\
& \texttt{user\_tweet\_count} & Total number of tweets posted by the user. \\
& \texttt{user\_listed\_count} & Number of public lists the user appears in. \\
\hline

\end{tabular}
\end{adjustwidth}
\end{table}

Since the tweet content constitutes a key source of semantic information, we applied a \textbf{text preprocessing }pipeline prior to vectorization. The text was lowercased, and all URLs, mentions (@), hashtags (\#), numbers, and punctuation were removed. Tokenization was performed using \texttt{nltk.word\_tokenize}, and English stopwords were removed using \texttt{nltk.corpus.stopwords}. We then trained a \textbf{Word2Vec} model~\cite{mikolov_efficient_2013} using the \texttt{gensim} library~\cite{rehurek_software_2010} to learn distributed word representations. The training parameters were set as follows: \texttt{vector\_size=30} (each word is represented as a vector of 30 dimension), \texttt{window=5} (context of 5 words before and after of the target term), and \texttt{min\_count=2} (we excluded words with a less than two frequency). Each tweet was embedded as the average of the Word2Vec vectors of its tokens. In cases where none of the words were found in the model's vocabulary, a zero vector of 30 dimensions was assigned. The resulting features (\texttt{w2v\_0} to \texttt{w2v\_29}) were concatenated with the remaining numeric and categorical features presented in\textbf{~\autoref{tab:features}}.

All numeric variables were \textbf{standardized} to ensure comparability across feature scales. We computed the Pearson correlation matrix and identified highly correlated feature pairs (absolute correlation > 0.8). Among the engagement metrics, only one of the variables was retained. Similarly, only one variable was kept between \texttt{user\_followers} and \texttt{user\_listed\_count}. However, we retained the substance mention indicators, such as \texttt{mentions\_ghb} and \texttt{mentions\_ecstasy}, despite their high negative correlation, as they capture distinct drug-specific information.

The result of this stage was a clean, vectorized, and standardized dataset, composed of linguistic, geographic, user, and engagement features, suitable for downstream classification modeling.

\subsection{Machine Learning modeling} \label{sec:ml_modeling}

Once the dataset was fully processed, we proceeded with the training and evaluation of several machine learning classifiers. Given the imbalanced nature of the data, where tweets labeled as negative ($0$) were significantly more prevalent than positive ($1$) ones, careful attention was paid to class imbalance to avoid bias and ensure robust learning. The final dataset was split into training (80\%) and test (20\%) subsets using a stratified split to preserve class proportions across both subsets. 

We implemented and compared a \textbf{diverse set of classification algorithms}, ranging from interpretable models to more complex ensemble and neural architectures. Specifically, we evaluated Decision Tree (DT)~\cite{quinlan_induction_1986} and Logistic Regression (LR)~\cite{cox_regression_1958} as baseline models, ensemble methods including Random Forest (RF)~\cite{breiman_random_2001}, Bagging~\cite{breiman_bagging_1996}, Adaptive Boosting (AdaBoost)~\cite{freund_decision-theoretic_1997}, and Extreme Gradient Boosting (XGBoost)~\cite{chen_xgboost_2016}, as well as a Multi-layer Perceptron (MLP)~\cite{rumelhart_learning_1986} to capture non-linear patterns in the data. These classifiers offer a broad spectrum of modeling capabilities in terms of interpretability, robustness, and predictive performance.

We tuned each model’s \textbf{hyperparameters} using randomized search over tailored parameter distributions, optimizing for F1-score due to its suitability for imbalanced classification. Each algorithm had a customized search space: for DT and RF, we varied splitting criteria, number of estimators, and feature selection strategies; Bagging explored sampling fractions, estimator count, and warm starts; AdaBoost varied learning rates and estimators; LR tested regularization types and compatible solvers; XGBoost was tuned over tree depth and estimator count; and MLP explored activation functions, hidden layer sizes, and learning rate strategies.

To further improve reliability and prevent overfitting during hyperparameter tuning, we adopted a nested, stratified 5-fold\textbf{ Cross Validation} (CV) strategy. The inner loop was used to select the best hyperparameters via random search, while the outer loop evaluated the generalization performance. 
% This approach provides a more honest estimate of a model's ability to generalize to unseen data and avoids optimistic bias.

To \textbf{evaluate} models' performance, we compute True Positives (TP), False Positives (FP), True Negatives (TN), and False Negatives (FN). During cross-validation, we report precision, recall, accuracy, and F1-score. For the final test evaluation, we additionally include the Area Under the Receiver Operating Characteristic Curve (AUROC) and the Area Under the Precision-Recall Curve (AUPRC), which provide threshold-independent assessments of discriminative performance, especially relevant in imbalanced settings.

% For assessing each model's performance, several complementary metrics were computed for both CV and final test evaluation:

% \begin{itemize}
%     \item \textbf{Confusion Matrix}: Summarizes predictions by showing True Positives (TP), False Positives (FP), True Negatives (TN), and False Negatives (FN). From these values, all the following metrics are derived.

%     \item \textbf{Recall}: Measures the proportion of actual positives that were correctly classified.
%     \begin{equation*}
%         \text{Recall} = \frac{TP}{TP + FN}
%     \end{equation*}

%     \item \textbf{Precision}: Measures the proportion of predicted positives that are actually correct.
%     \begin{equation*}
%         \text{Precision} = \frac{TP}{TP + FP}
%     \end{equation*}

%     \item \textbf{Accuracy}: Overall correctness of predictions; may be misleading in imbalanced settings.
%     \begin{equation*}
%         \text{Accuracy} = \frac{TP + TN}{TP + TN + FP + FN}
%     \end{equation*}

%     \item \textbf{F1-score}: Harmonic mean of precision and recall; robust for imbalanced classification.
%     \begin{equation*}
%         \text{F1-score} = 2 \cdot \frac{\text{Precision} \cdot \text{Recall}}{\text{Precision} + \text{Recall}}
%     \end{equation*}

%     \item Area Under the Receiver Operating Characteristic Curve (\textbf{AUROC}): Reflects the model's ability to distinguish between the two classes across various thresholds.

%     \item Area Under the Precision-Recall Curve (\textbf{AUPRC}): Especially informative under severe imbalance, focusing on how well the model identifies the minority class.
% \end{itemize}

As the positive class (tweets indicating \textit{desired} drug-related effects) was significantly underrepresented, we compared three distinct \textbf{strategies for handling class imbalance}:

\begin{itemize}
    \item \textbf{No sampling (baseline)}: The original distribution of positive and negative tweets was maintained, which served as a reference to assess the effectiveness of imbalance mitigation techniques.

    \item \textbf{Cost-sensitive learning}: For classifiers that support weighted training, we applied \texttt{class\_weight=`balanced'}. This approach adjusts the loss function to penalize misclassifications of the minority class more heavily, encouraging the model to give more attention to underrepresented instances without modifying the training data.

    \item \textbf{Oversampling}. We applied Synthetic Minority Over-sampling Technique (SMOTE) to synthetically generate new positive-class samples by interpolating between existing minority-class examples in the feature space~\cite{chawla_smote_2002}. Two configurations were tested: (i) \textbf{\textit{Pre-CV SMOTE}}, where oversampling was applied once to the entire training set before starting the CV process (computationally simpler, but it carries a risk of information leakage, as synthetic samples derived from the training data may indirectly influence the validation folds, leading to overoptimistic performance estimates); and (ii) \textbf{\textit{In-CV SMOTE}}, where SMOTE was applied within each fold of the CV pipeline (for each train/validation split, oversampling was performed only on the training portion of the fold, and the synthetic samples were never present in the validation set, ensuring a more realistic evaluation by preventing data leakage and closely simulating the generalization capability of the model in unseen data). In both setups, SMOTE was always applied exclusively to the training data and never to the test set.
\end{itemize}

\section{Results and discussion} \label{sec:results}

We began our study by collecting a total of 6,755,394 tweets from 17 JSON archives, authored by 2,681,817 unique users. The first filtering step involved identifying whether the tweet mentioned any of the three target substances (ecstasy, GHB, or 2C-B). This filtering yielded 5,228,085 tweets (77.4\% of the total), which became the foundation of our analysis. Within this subset, most tweets (5,212,224) referenced only one substance, while a small fraction (15,828 tweets) mentioned two, and an even smaller portion (33 tweets) mentioned three or more substances simultaneously.

\begin{figure}[h]
\begin{adjustwidth}{-3.8cm}{-3.8cm}
\centering
\includegraphics[width=1.6\textwidth]{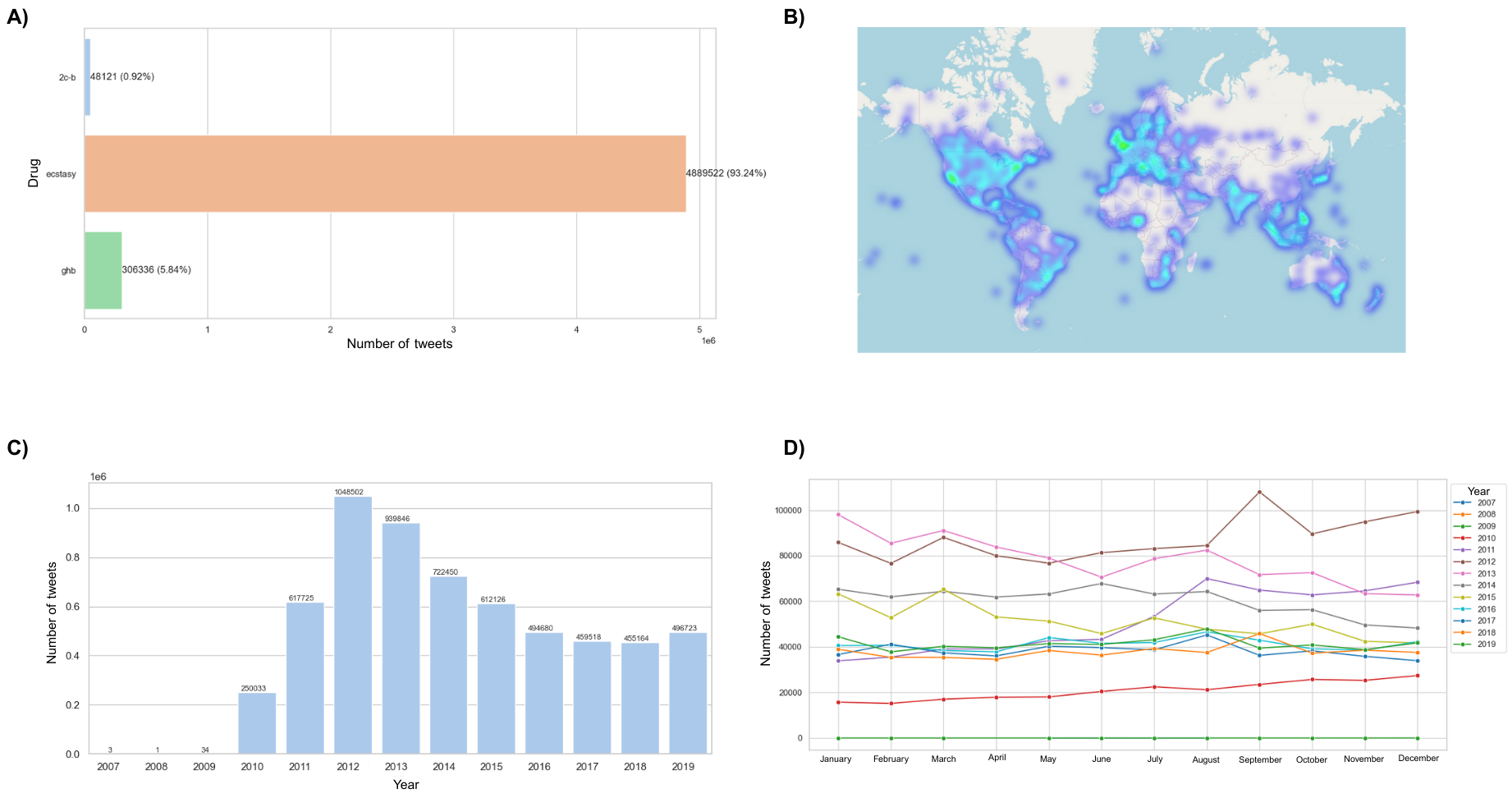}
\caption{General descriptive analysis of the dataset. A) Number of tweets per drug. B) Heatmap showing number of tweets per geolocation. C) Number of tweets per year. D) Number of tweets per month and year.}
\label{fig:descriptive_analysis}
\end{adjustwidth}
\end{figure}

\textbf{\autoref{fig:descriptive_analysis}} provides an overview of this dataset descriptive analysis. The distribution of substances revealed a strong predominance of ecstasy-related content (4,889,522 tweets), followed by GHB (306,336) and 2C-B (48,121), reflecting broader usage or cultural popularity in social discourse. In terms of associated terms, slang terms were identified in 1,104,696 tweets, whereas MetaMap terms were associated to 4,457,981 tweets.

From a metadata perspective, 420,731 tweets included multimedia content, and 166,765 tweets contained geolocation metadata. Regarding user characteristics, 49,421 users (1.8\%) were verified accounts, contributing to 2,593,637 tweets, while the rest were from unverified profiles.

To proceed with the construction of the classification dataset, we applied an annotation pipeline leveraging weak supervision. We began with the pool of tweets containing drug mentions and associated either to slang or MetaMap terms, totaling 5,562,677 tweets. Each term identified in those tweets had been previously annotated by three domain experts with a polarity label: \emph{positive}, \emph{negative}, or \emph{context-dependent}. Only tweets in which a majority consensus (i.e., at least 2 of 3 annotators) was reached for all annotated terms were retained, resulting in 2,986,159 tweets.

From these, we separated tweets into those annotated using slang terms (345,422) and those annotated via MetaMap (1,225,003), depending on the original matching source. Tweets labeled as context-dependent were discarded (1,263,910), as they required nuanced interpretation beyond automated heuristics. After merging both annotation streams and consolidating duplicate tweets, we obtained a final classification dataset containing 92,291 tweets, each labeled as either expressing positive or negative polarity with respect to recreational drug effects.

The final classification dataset exhibits a strong class imbalance, with only 4,237 tweets (4.59\%) labeled as positive, indicating references to personal or experiential effects of drug use, and 88,352 tweets (95.41\%) labeled as negative. When examining the distribution by substance, this imbalance is even more pronounced in certain categories. For ecstasy, the most prevalent substance in the dataset, there are 4,163 positive tweets compared to 81,937 negatives. In the case of GHB, only 59 tweets were labeled as positive versus 5,974 negatives, while 2C-B has 15 positive tweets and 441 negatives. This imbalance (particularly for GHB and 2C-B) reflects the lower frequency of explicit, experiential positive mentions of these substances in user-generated content.

Moving to the ML modeling,\textbf{~\autoref{tab:cv_results}} summarizes the average performance and standard deviation of each model across CV folds, evaluated under four different balancing strategies: no sampling, cost-sensitive learning, SMOTE applied before CV (pre-CV), and SMOTE applied within the folds (in-CV).

\begin{table}[!ht]
\centering
\caption{Mean and standard deviation of CV metrics for each ML method and balancing strategy.}
\label{tab:cv_results}
\begin{tabular}{|c|c|c|c|c|c|c|}
\hline
\textbf{Balancing} & \textbf{Method} & \textbf{F1 score} & \textbf{Recall} & \textbf{Precision} & \textbf{Accuracy} \\
\hline
\multirow{7}{*}{\shortstack{No\\sampling}} 
& AdaBoost & 0.786 ± 0.025 & 0.980 ± 0.002 & 0.725 ± 0.024 & 0.857 ± 0.025 \\
& Bagging & 0.849 ± 0.013 & 0.986 ± 0.001 & 0.754 ± 0.019 & 0.971 ± 0.006 \\
& DT & 0.727 ± 0.009 & 0.973 ± 0.002 & 0.726 ± 0.020 & 0.744 ± 0.010 \\
& LR & 0.815 ± 0.006 & 0.983 ± 0.001 & 0.745 ± 0.009 & 0.898 ± 0.004 \\
& MLP & 0.874 ± 0.006 & 0.988 ± 0.001 & \textbf{0.836 ± 0.013} & 0.923 ± 0.010 \\
& RF & 0.845 ± 0.015 & 0.986 ± 0.001 & 0.736 ± 0.019 & \textbf{0.986 ± 0.008} \\
& XGBoost & \textbf{0.887 ± 0.012} & \textbf{0.989 ± 0.001} & 0.829 ± 0.019 & 0.953 ± 0.003 \\
\hline
\multirow{7}{*}{\shortstack{Cost\\sensitive}} 
& AdaBoost & 0.795 ± 0.018 & 0.981 ± 0.002 & 0.867 ± 0.027 & 0.735 ± 0.024 \\
& Bagging & 0.852 ± 0.014 & 0.986 ± 0.001 & 0.968 ± 0.012 & 0.758 ± 0.021 \\
& DT & 0.733 ± 0.020 & 0.976 ± 0.002 & 0.753 ± 0.015 & 0.745 ± 0.024 \\
& LR & 0.581 ± 0.004 & 0.931 ± 0.001 & 0.420 ± 0.003 & \textbf{0.942 ± 0.012} \\
& MLP & 0.883 ± 0.010 & 0.989 ± 0.001 & 0.930 ± 0.015 & 0.852 ± 0.017 \\
& RF & 0.842 ± 0.009 & 0.986 ± 0.001 & \textbf{0.987 ± 0.008} & 0.739 ± 0.016 \\
& XGBoost &\textbf{ 0.894 ± 0.013} & \textbf{0.990 ± 0.001} & 0.931 ± 0.012 & 0.860 ± 0.020 \\
\hline
\multirow{7}{*}{\shortstack{SMOTE\\(pre-CV)}} 
& AdaBoost & 0.949 ± 0.001 & 0.948 ± 0.001 & 0.943 ± 0.002 & 0.954 ± 0.003 \\
& Bagging & \textbf{0.996 ± 0.000} & \textbf{0.996 ± 0.000} & \textbf{0.994 ± 0.001} & 0.998 ± 0.000 \\
& DT & 0.970 ± 0.001 & 0.970 ± 0.001 & 0.964 ± 0.001 & 0.976 ± 0.001 \\
& LR & 0.953 ± 0.002 & 0.953 ± 0.002 & 0.949 ± 0.006 & 0.957 ± 0.007 \\
& MLP & 0.995 ± 0.000 & 0.995 ± 0.000 & 0.990 ± 0.001 & \textbf{0.999 ± 0.000} \\
& RF & 0.993 ± 0.000 & 0.993 ± 0.001 & 0.991 ± 0.001 & 0.996 ± 0.000 \\
& XGBoost & 0.990 ± 0.001 & 0.990 ± 0.001 & 0.986 ± 0.000 & 0.994 ± 0.000 \\
\hline
\multirow{7}{*}{\shortstack{SMOTE\\(in-CV)}} 
& AdaBoost & 0.622 ± 0.011 & 0.943 ± 0.002 & 0.469 ± 0.008 & \textbf{0.925 ± 0.018} \\
& Bagging & \textbf{0.886 ± 0.015} & \textbf{0.988 ± 0.002} & \textbf{0.897 ± 0.018} & 0.876 ± 0.020 \\
& DT & 0.672 ± 0.010 & 0.957 ± 0.001 & 0.569 ± 0.015 & 0.821 ± 0.022 \\
& LR & 0.644 ± 0.022 & 0.949 ± 0.005 & 0.502 ± 0.026 & 0.898 ± 0.014 \\
& MLP & 0.861 ± 0.010 & 0.986 ± 0.001 & 0.852 ± 0.021 & 0.871 ± 0.013 \\
& RF & 0.856 ± 0.009 & 0.985 ± 0.001 & 0.848 ± 0.012 & 0.866 ± 0.022 \\
& XGBoost & 0.828 ± 0.007 & 0.982 ± 0.001 & 0.795 ± 0.013 & 0.859 ± 0.015 \\
\hline
\end{tabular}
\end{table}

Overall, the results show a clear benefit of using data balancing, with SMOTE (pre-CV) yielding the best results across most models. Under this setting, Bagging, MLP, and RF achieved F1-scores above 0.99, alongside very high AUROC and AUPRC values, indicating near-perfect separability between classes. XGBoost also maintained strong performance ($F1 = 0.990 \pm 0.001$), confirming its robustness in this setting. However, these nearly perfect metrics (combined with zero standard deviation across folds in some cases) suggest signs of overfitting, likely due to data leakage introduced by applying SMOTE before the CV split. This leads to synthetic examples appearing in both train and validation folds, inflating performance metrics unrealistically.

To address this issue, we also tested SMOTE applied within each fold of the CV (in-CV), thus ensuring that oversampling was performed only on the training data in each split. In this more realistic setting, MLP and RF again stood out ($F1 = 0.861 \pm 0.010$ and $0.856 \pm 0.009$, respectively), followed by XGBoost and Bagging, all of which maintained strong performance with better generalization and more meaningful standard deviations. The cost-sensitive strategy also performed strongly, particularly with XGBoost and MLP ($F1 = 0.894 \pm 0.013$ and $0.883 \pm 0.010$), although Logistic Regression suffered a notable drop in precision and F1-score. This was expected given the model’s linear nature and sensitivity to class imbalance, even with weighting. In the no sampling configuration, the best results were obtained again with XGBoost and MLP, followed by RF and LR. Although precision and recall were reasonably balanced in this setting, F1-scores were consistently lower than those achieved with any of the balancing strategies.

% In summary, these results highlight that balancing strategies, especially applied correctly within CV folds, or cost-sensitive learning, markedly improve performance, particularly when combined with robust classifiers such as MLP, RF, and XGBoost. Moreover, reporting standard deviation across folds proves essential to detect signs of overfitting and assess the stability and generalizability of models.

\textbf{~\autoref{tab:test_results}} shows the classification performance of each ML model on the held-out test set across all balancing strategies. As expected from the CV phase, the results reveal that the best performing models in terms of F1-score were again XGBoost and MLP, particularly when using SMOTE either before or within CV. The highest F1-score was achieved by XGBoost trained with SMOTE during CV ($F1 = 0.8668$), followed closely by MLP ($F1 = 0.8476$) and RF ($F1 = 0.8533$) under the same strategy. These models also reported strong recall and precision trade-offs, with AUROC and AUPRC consistently above 0.91.

\begin{table}[ht]
\begin{adjustwidth}{-3cm}{-3cm} 
\centering
\caption{Performance on the test set for each ML method and balancing strategy.}
\label{tab:test_results}
% \scriptsize
\begin{tabular}{|c|c|c|c|c|c|c|c|c|c|c|c|}
\hline
\textbf{Balancing} & \textbf{Method} & \textbf{F1} & \textbf{Recall} & \textbf{Precision} & \textbf{Accuracy} & \textbf{TN} & \textbf{FN} & \textbf{FP} & \textbf{TP} & \textbf{AUROC} & \textbf{AUPRC} \\
\hline
\multirow{7}{*}{\shortstack{No\\sampling}} 
& AdaBoost & 0.7869 & 0.7344 & 0.8475 & 0.9797 & 12966 & 187 & 93 & 517 & 0.9769 & 0.8480 \\
& Bagging & 0.8374 & 0.7386 & 0.9665 & 0.9853 & 13041 & 184 & 18 & 520 & 0.9720 & 0.8938 \\
& DT & 0.7408 & 0.7472 & 0.7346 & 0.9733 & 12869 & 178 & 190 & 526 & 0.8663 & 0.5618 \\
& LR & 0.8158 & 0.7486 & 0.8963 & 0.9827 & 12998 & 177 & 61 & 527 & 0.9817 & 0.8825 \\
& MLP & 0.8536 & \textbf{0.8196} & 0.8904 & 0.9856 & 12988 & \textbf{127} & 71 & 577 & 0.9889 & 0.9219 \\
& RF & 0.8430 & 0.7358 & \textbf{0.9867} & 0.9860 & 13052 & 186 & \textbf{7} & 518 & 0.9874 & 0.9187 \\
& XGBoost & \textbf{0.8807} & 0.8182 & 0.9536 & \textbf{0.9887} & \textbf{13031} & 128 & 28 & \textbf{576} & \textbf{0.9895} & \textbf{0.9309} \\
\hline
\multirow{7}{*}{\shortstack{Cost\\sensitive}} 
& AdaBoost & 0.7869 & 0.7344 & 0.8475 & 0.9797 & 12966 & 187 & 93 & 517 & 0.9769 & 0.8480 \\
& Bagging & 0.8429 & 0.7472 & 0.9669 & 0.9858 & 13041 & 178 & 18 & 526 & 0.9688 & 0.8906 \\
& DT & 0.7446 & 0.7287 & 0.7611 & 0.9744 & 12898 & 191 & 161 & 513 & 0.8582 & 0.5685 \\
& LR & 0.5750 & \textbf{0.9418} & 0.4139 & 0.9288 & 12120 & \textbf{41} & 939 & \textbf{663} & 0.9829 & 0.8525 \\
& MLP & 0.8724 & 0.8452 & 0.9015 & 0.9874 & 12994 & 109 & 65 & 595 & \textbf{0.9895} & 0.9311 \\
& RF & 0.8392 & 0.7301 & \textbf{0.9866} & 0.9857 & \textbf{13052} & 190 & \textbf{7} & 514 & 0.9872 & 0.9155 \\
& XGBoost & \textbf{0.8852} & 0.8438 & 0.9310 & \textbf{0.9888} & 13015 & 110 & 44 & 594 & 0.9890 & \textbf{0.9339} \\
\hline
\multirow{7}{*}{\shortstack{SMOTE\\(pre-CV)}} 
& AdaBoost & 0.5980 & 0.8778 & 0.4534 & 0.9396 & 12314 & 86 & 745 & 618 & 0.9740 & 0.8254 \\
& Bagging & 0.8280 & 0.8480 & 0.8089 & 0.9820 & 12918 & 107 & 141 & 597 & 0.9856 & 0.9066 \\
& DT & 0.6854 & 0.8125 & 0.5927 & 0.9619 & 12666 & 132 & 393 & 572 & 0.8912 & 0.4912 \\
& LR & 0.6105 & \textbf{0.9219} & 0.4564 & 0.9398 & 12286 & \textbf{55} & 773 & \textbf{649} & 0.9824 & 0.8534 \\
& MLP & 0.8438 & 0.8480 & 0.8397 & 0.9839 & 12945 & 107 & 114 & 597 & 0.9872 & 0.9234 \\
& RF & 0.8498 & 0.8480 & 0.8516 & 0.9847 & 12955 & 107 & 104 & 597 & \textbf{0.9886} & 0.9205 \\
& XGBoost & \textbf{0.8721} & 0.8523 & \textbf{0.8929} & \textbf{0.9872} & \textbf{12987} & 104 & \textbf{72} & 600 & 0.9881 & \textbf{0.9308} \\
\hline
\multirow{7}{*}{\shortstack{SMOTE\\(in-CV)}} 
& AdaBoost & 0.6355 & 0.8793 & 0.4976 & 0.9484 & 12434 & 85 & 625 & 619 & 0.9767 & 0.8430 \\
& Bagging & 0.8151 & 0.8452 & 0.7870 & 0.9804 & 12898 & 109 & 161 & 595 & 0.9834 & 0.8980 \\
& DT & 0.6809 & 0.8153 & 0.5845 & 0.9609 & 12651 & 130 & 408 & 574 & 0.8920 & 0.4860 \\
& LR & 0.6091 & \textbf{0.9219} & 0.4548 & 0.9395 & 12281 & \textbf{55} & 778 & \textbf{649} & 0.9824 & 0.8537 \\
& MLP & 0.8476 & 0.8452 & 0.8500 & 0.9845 & 12954 & 109 & 105 & 595 & 0.9876 & 0.9184 \\
& RF & 0.8533 & 0.8509 & 0.8557 & 0.9850 & 12958 & 105 & 101 & 599 & 0.9885 & 0.9194 \\
& XGBoost & \textbf{0.8668} & 0.8551 & \textbf{0.8788} & \textbf{0.9866} & \textbf{12976} & 102 & \textbf{83} & 602 & \textbf{0.9894} & \textbf{0.9325} \\
\hline
\end{tabular}
\end{adjustwidth}
\end{table}

In contrast, LR exhibited limited effectiveness across most settings, especially in the cost-sensitive and SMOTE-based configurations, with a notable drop in precision and accuracy. This indicates that LR may not capture the complex patterns in the data as efficiently as ensemble or neural approaches. The cost-sensitive strategy offered competitive results, particularly for MLP and XGBoost, although in most cases, SMOTE-based approaches led to superior precision-recall balance. Models trained with SMOTE before CV achieved strong recall but tended to suffer from lower precision, especially in methods like AdaBoost and LR, indicating a higher rate of false positives. On the other hand, SMOTE applied within each fold (in-CV) proved to be a more robust strategy.

\section{Conclusions and future works} \label{sec:conclusion}

In this study, we presented a comprehensive pipeline for analyzing recreational drug use discourse on Twitter, with a focus on three substances: ecstasy, GHB, and 2C-B. Starting from over 6.7 million tweets, we developed a domain-informed annotation strategy combining weak supervision, expert-curated polarity labels, and semantic matching through slang and MetaMap terms. This process led to the creation of a polarity-labeled dataset of over 92,000 tweets, capturing both positive and negative mentions of drug-related experiences. We evaluated multiple ML classifiers under different strategies to mitigate class imbalance, including cost-sensitive learning and SMOTE-based oversampling. Our results demonstrate that appropriate balancing, particularly SMOTE applied within each fold of CV, significantly enhances classification performance. Ensemble methods such as XGBoost, RF, and Bagging, along with neural networks, consistently outperformed simpler models like LR. The highest test F1-score was obtained using XGBoost with in-fold SMOTE, reaching 0.8668 and supported by strong precision, recall, AUROC, and AUPRC values.

Despite these promising results, several challenges remain. The strong class imbalance in real-world data, particularly for less frequently mentioned substances like GHB and 2C-B, limits the generalizability of the models. Additionally, the reliance on weak supervision and term-level annotations, while scalable, may miss subtleties in language use and context.

In future work, we plan to expand our annotation framework by incorporating large language models to refine polarity inference and disambiguate context-dependent mentions. We also aim to explore further temporal and geographical patterns in user discourse, as well as the interplay between drug mentions and mental health indicators. Finally, we envision extending this approach to other substances and social platforms to build a broader, cross-source understanding of public perceptions and self-reported experiences related to drug use.

\subsection*{Acknowledgments}
We would like to acknowledge the contributions of Alejandro Gouloumis Contreras, whose Master's Thesis~\cite{goulomis_tfm_2022} laid the groundwork for the extraction and initial processing of the Twitter data used in this study.

\bibliographystyle{splncs04}
\bibliography{references}  
\end{document}